\documentclass[11pt,a4paper]{article}
\usepackage{anyfontsize}
\usepackage[hyperref]{emnlp2018}
\usepackage{times}
\usepackage{latexsym}
\usepackage{color}
\usepackage{url}
\usepackage[draft,textsize=footnotesize]{todonotes}
\usepackage{ulem}
\usepackage{multirow}
\usepackage[capitalize]{cleveref}
\usepackage{enumitem}
\usepackage{tabularx}

\aclfinalcopy       
\title{How do you correct run-on sentences it's not as easy as it seems}

\author{Junchao Zheng, Courtney Napoles, Joel Tetreault \and Kostiantyn Omelianchuk\\
Grammarly\\
{\tt firstname.lastname@grammarly.com}}
\date{}

\begin{document}
\maketitle
\begin{abstract}
Run-on sentences are common grammatical mistakes %made by native and non-native writers
but little research has tackled this problem to date.  This work introduces two machine learning models to correct run-on sentences that outperform
% and evaluate them against 
leading methods for related tasks, punctuation restoration and whole-sentence grammatical error correction.
% We show that our models outperform these other methods. 
Due to the limited annotated data for this error, we experiment with artificially generating training data from clean newswire text. Our findings suggest artificial training data is viable for this task.
% run-ons  performance degrades at least 10\% on run-ons found in text containing other grammatical errors compared to text that has no other mistakes. 
We discuss implications for correcting run-ons and other types of mistakes that have low coverage in error-annotated corpora.
\end{abstract}

\section{Introduction}

A run-on sentence is defined as having at least two main or independent clauses that lack either a conjunction to connect them or a punctuation mark to separate them. Run-ons are problematic because they not only make the sentence unfriendly to the reader but potentially also to the local discourse. Consider the example in Table~\ref{Tab:1}.

In the field of grammatical error correction (GEC), most work has typically focused on determiner, preposition, verb and other errors which non-native writers make more frequently.  Run-ons have received little to no attention even though they are common errors for both native and non-native speakers. Among college students in the United States, run-on sentences are the 18th most frequent error and the 8th most frequent error made by students who are not native English speakers~\cite{leacock2014automated}.

\begin{table}[t]
\begin{center}
\fontsize{10}{13}
\selectfont
\begin{tabularx}{\linewidth}{|X|}
\hline
\textit{Before correction} \\
But the illiterate will not stay illiterate always if they put an effort to improve and are given a chance for good education, they can still develop into a group of productive Singaporeans. \\ \\
\hline 
\textit{After correction} \\ 
But the illiterate will not stay illiterate always. If they put an effort to improve and are given a chance for good education, they can still develop into a group of productive Singaporeans. \\ \\
\hline 
\end{tabularx}
\end{center}
\caption{A run-on sentence before and after correction.}
\label{Tab:1} 
\end{table}

Correcting run-on sentences is challenging \cite{kagan1980run} for several reasons:
\begin{itemize}[noitemsep]
\item They are sentence-level mistakes with long-distance dependencies, whereas most other grammatical errors are local and only need a small window for decent accuracy.
\item There are multiple ways to fix a run-on sentence. For example, one can a) add sentence-ending punctuation to separate them; b) add a conjunction (such as \textit{and}) to connect the two clauses; or c) convert an independent clause into a dependent clause. \\
\item They are relatively infrequent in existing, annotated GEC corpora and therefore existing systems tend not to learn how to correct them.
\end{itemize}
% In this initial exploration, we consider only the first scenario and predict missing periods.

%In this paper, we correct run-on sentences by simply inserting a {\tt PERIOD} or replacing a comma splice with a {\tt PERIOD}. 

%JRT - not really sure what this sentence means: Though there are several other options, in this way we can cover all kinds of run-on sentence correction in one theme. 
%We identify deficiencies of leading approaches to GEC and punctuation restoration, which either do not correct any run-on sentences in our experiments or have very low precision on noisy real-world text.  
% \todo{JRT: do we actually do this?  it makes it sound like we find flaws in the algo, but it's probably likely the training data right?  So maybe "flaws" should be "deficienceies"? CN: noted}  
%, respectively.
% that leading approaches to GEC does not correct run-on sentences and a punctuation restoration do not correct run-on sentences well.
In this paper, we analyze the task of automatically correcting run-on sentences.  We develop two methods: a conditional random field model (roCRF) and a Seq2Seq attention model (roS2S) and show that they outperform models from the sister tasks of punctuation restoration and whole-sentence grammatical error correction.  
%Both models work correct a run-on by simply inserting a {\tt PERIOD} or replacing a comma splice with a {\tt PERIOD}. % CN: I moved (seq2seq) to related work because of the page layout
% and show that they outperform models from the sister task of punctuation restoration and whole-sentence grammatical error correction.  
We also experiment with artificially generating training examples in clean, otherwise grammatical text, and show that models trained on this data do nearly as well predicting artificial and naturally occurring run-on sentences.

\section{Related Work}

Early work in the field of GEC focused on correcting specific error types such as preposition and article errors~\cite{tetreault-foster-chodorow:2010:Short,rozovskaya-roth:2011:ACL-HLT2011,dahlmeier-ng:2011:ACL-HLT2011}, but did not consider run-on sentences.  The closest work to our own is~\citet{israel-tetreault-chodorow:2012:NAACL-HLT}, who used Conditional Random Fields (CRFs) for correcting comma errors (excluding comma splices, a type of run-on sentence). ~\citet{Lee-EtAl:2014:PACLIC} used a similar system based on CRFs but focused on comma splice correction. %, achieving an average of 89\% precision and 25\% recall on essays from both native and non-native English learners. 
% However, the system is not capable of dealing with \textit{comma splices}, a common type of run-on sentence.\todo{@JZ: how did you treat comma splices in NUCLE? JZ: I annotated manually on each comma splices with the Srun tag. Example: line 28-30 in NUCLE\_sent2review}
Recently, the field has focused on the task of \textit{whole-sentence correction}, targeting all errors in a sentence in one pass.
%using phrase-based and neural machine translation techniques~\cite{Cho2014,Yuan2016}. \todo{remove }
Whole-sentence correction methods borrow from advances in statistical machine translation~\cite{W12-2005,felice-EtAl:2014:W14-17,junczysdowmunt-grundkiewicz:2016:EMNLP2016} and, more recently, neural machine translation~\cite{yuan-briscoe:2016:N16-1,chollampatt2018multilayer,xie-EtAl:2018:N18-1,junczysdowmunt-EtAl:2018:N18-1}.
%,grundkiewicz-junczysdowmunt:2018:N18-2}. 
% For example, ~\citet{junczys2016phrase} investigated interactions of dense and sparse features and so on to optimize \(M^{2}\) score while ~\citet{napoles2017jfleg} suggested NMT can return more fluent output. ~\citet{grundkiewicz2018near} incorporated SMT and NMT to preserve both accuracy and fluency to get closer to reaching human-level performance.

To date, GEC systems have been evaluated on corpora of non-native student writing such as NUCLE~\cite{dahlmeier-ng-wu:2013:BEA8} and the Cambridge Learner Corpus First Certificate of English~\cite{yannakoudakis-briscoe-medlock:2011:ACL-HLT2011}.
The 2013 and 2014 CoNLL Shared Tasks in GEC used NUCLE as their train and test sets~\cite{ng-EtAl:2013:CoNLLST,ng-EtAl:2014:W14-17}. There are few instances of run-on sentences annotated in both test sets, making it hard to assess system performance on that error type.
%\todoz{There are few instances of run-on sentences annotated in both test sets, making it hard to assess system performance on that error type. ==> According to the annotation, there are few instances of run-on sentences in both test sets, making it hard to assess system performance on that error type. -BP7}
% In \S\ref{sec:data} we discuss details about our training and test sets to address this problem.\todo{JRT: we could also just get rid of this sentence}

A closely related task to run-on error correction is that of \textit{punctuation restoration} in the automatic speech recognition (ASR) field.  Here, a system takes as input a speech transcription and is tasked with inserting any type of punctuation where appropriate.
%Work in the automatic speech recognition (ASR) community 
%A closely related task in automatic speech recognition (ASR) systems is \textit{punctuation restoration}. 
Most work utilizes textual features with n-gram models~\cite{gravano2009restoring}, CRFs~\cite{lu-ng:2010:EMNLP}, convolutional neural networks or recurrent neural networks ~\cite{peitz2011modeling,che16punctuation}. The Punctuator~\cite{tilk16punctuator} is a leading punctuation restoration system based on a sequence-to-sequence model (Seq2Seq) %reaches an average of 70\% precision and 60\% recall restoring comma, period and question marks on speech transcripts~\cite{tilk16punctuator}. The Punctuator 
% predicts all missing punctuation and 
trained on long slices of text which can span multiple sentences.

\begin{table}[t!]
\begin{center}
% \small
% \begin{minipage}{15cm}
% \begin{tabular}{llllllllll }
% % \hline
% Accordingly&,&people&enjoy&and&better&health&with&a& longer\\
% \spc & \spc & \spc &\spc &\spc &\spc &\spc &\spc &\spc &\spc\\
% % \hline
% \end{tabular}

% \vspace{.2cm}\rule{15cm}{0.4pt}\vspace{.2cm}

% \begin{tabular}{lllllllllllll}
% % \hline
% age & This & enables & them & to & stay & in & work & for & a \\
%  \tt PERIOD & \spc &\spc &\spc &\spc &\spc &\spc &\spc &\spc &\spc \\
% % \hline 
% \end{tabular}

% \vspace{.2cm}\rule{15cm}{0.4pt}\vspace{.2cm}

% \begin{tabular}{lll}
% longer & time & .\\
% \spc & \spc & \spc\\
% \end{tabular}
% \end{minipage}

\fontsize{10}{13}
\selectfont
\begin{tabularx}{\linewidth}{|X|}
\hline
 This\verb|/S|  shows\verb|/S| the\verb|/S| rising\verb|/S| of\verb|/S| life\verb|/S| expectancies\verb|/P| it\verb|/S|  is\verb|/S| an\verb|/S| achievement\verb|/S| and\verb|/S| it\verb|/S| is\verb|/S| also\verb|/S| a\verb|/S| challenge\verb|/S| .\verb|/S|\\
\hline
\end{tabularx}
\end{center}
\caption{\label{tab:2} NUCLE sentence labeled to indicate what follows each token: a space ({\tt  S}) or period ({\tt P}).}
\end{table}

\section{Model Descriptions}

% In contrast to recent GEC systems that perform whole-sentence correction, we develop a dedicated model targeting run-on sentences due to their infrequency in annotated GEC corpora.
% \todo{JRT: is that the main reason we are doing an error-specific model?  Burden is on us to show that whole sentence methods don't perform well, right?  before saying this? CN: is it?}  
We treat correcting run-ons as a sequence labeling task: 
given a sentence, the model reads each token and learns whether there is a {\tt SPACE} or {\tt PERIOD} 
%\footnote{This work only targets missing periods because we have no examples of missing question marks or exclamation points in the existing corpora.} 
following that token, as shown in Table~\ref{tab:2}. 
We apply two sequence models to this task, conditional random fields (\textit{roCRF}) and Seq2Seq (\textit{roS2S}).

% \todoc{what labels are predicted? just missing punctuation or what punctuation is missing?}  \todot{agreed: we should clearly delineate the problem here in one paragraph}  \todot{list the approaches we take in this intro area.}
% \todo{start with the simpler models and go up.  So CRFs first}
% \todo{This section we will likely shrink since we aren't making any crazy new model ideas.  I do like the attention to detail and descriptions Junchao.}

\subsection{Conditional Random Fields}\label{sec:crf}
Our CRF model, roCRF, represents a sentence as a sequence of spaces between tokens, labeled to indicate whether a period should be inserted in that space. 
% model the task of run-on sentence correction as assigning a label to the space following each token in the sentence. 
% The label to each space is binary indicating whether a period is missing in that space. 
Each space is represented by contextual features (sequences of tokens, part-of-speech tags, and capitalization flags around each space), parse features %\todo{JRT: which parser? CN: I can add which parser in the supp. material but IDK what it was trained on.} 
(the highest uncommon ancestor of the word before and after the space, and binary indicators of whether the highest uncommon ancestors are preterminals), and a flag indicating whether the mean per-word perplexity of the text decreases when a period is inserted at the space according to a 5-gram language model. %\todo{JRT: which LM is used? CN: common crawl but per DimaT we can't say we used that...} 

\subsection{Sequence to Sequence Model with Attention Mechanism}

%We use a Seq2Seq model as the baseline model. %CN - not baseline
Another approach is to treat it as a form of neural sequence generation. In this case, the input sentence is a single run-on sentence. During decoding we pass the binary label which determines if there is terminal punctuation following the token at the current position. We then combine the generated label and the input sequence to get the final output. 

Our model, roS2S, is a Seq2Seq attention model based on the neural machine translation model~\cite{bahdanau14neural}. The encoder is a bidirectional LSTM, where a recurrent layer processes the input sequence in both forward and backward direction. The decoder is a uni-directional LSTM. An attention mechanism is used to obtain the context vector.

\section{Data}\label{sec:data}

\paragraph{Train:} 

Although run-on sentences are common mistakes, existing GEC corpora do not include enough labeled run-on sentences to use as training data. 
Therefore we artificially generate training examples from a corpus of clean newswire text, Annotated Gigaword~\cite{napoles-gormley-vandurme:2012:AKBC-WEKEX}.
% We choose clean text as a source for artificially generating data because we are fairly confident that it does not contain any run-on sentences which could be present in corpora of noisy text.
We randomly select paragraphs and identify candidate pairs of adjacent sentences, where the sentences have between 5--50 tokens and no URLs or special punctuation (colons, semicolons, dashes, or ellipses).
Run-on sentences are generated by removing the terminal punctuation between the sentence pairs and lowercasing the first word of the second sentence (when not a proper noun).
In total we create 2.8 million run-on sentences, and randomly select 1.75M other Gigaword sentences for negative examples. We want the model to learn more patterns of run-on errors by feeding a large portion of positive examples while we report our results on a test set where the ratio is closer to that of real world. We call this data \textit{FakeGiga-Train}.
An additional 28k run-ons and 218k non-run-ons are used for validation.

\paragraph{Test:}

We evaluate on two dimensions: clean versus noisy text
%with artificial run-ons, 
and real versus artificial run-ons.
%in ESL text.
In the first evaluation, we artificially generate sentences from Gigaword and NUCLE following the procedure above such that 10\% of sentences are run-ons, based on our estimates of their rate in real-world data (similar observations can be found in \citet{watcharapunyawong2012thai}). We refer to these test sets as \textit{FakeGiga} and \textit{FakeESL} respectively. Please note that the actual run-on sentences in NUCLE are not included in \textit{FakeESL}. 

The second evaluation compares performance on artificial versus naturally occurring run-on sentences, using the NUCLE and CoNLL 2013 and 2014 corpora.
Errors in these corpora are annotated with corrected text and error types, one of which is \textit{Srun}: run-on sentences and comma splices.  \textit{Sruns} occur 873 times in the NUCLE corpus. 
We found that some of the \textit{Srun} annotations do not actually correct run-on sentences, so we reviewed the \textit{Srun} annotations to exclude any corrections that do not address run-on sentences. We also found that there are 300 out of the 873 sentences with \textit{Srun} annotations which actually perform correction by adding a period. 
Other \textit{Srun} annotations correct run-ons by converting independent clauses to dependent clauses, but we only target missing periods in this initial work.
We manually edit \textit{Srun} annotations so that the only correction performed is inserting periods. (This could be as simple as deleting the comma in the original text of a comma splice or more involved, as in rewriting a dependent clause to an independent clause in the corrected text.) 
% Examples are included in the Supplementary Material.)
In total, we find fewer than 500 run-on sentences.
Running the same procedure over the CoNLL 2013 and 2014 Shared Task test sets results in 59 more run-on sentences and 2,637 non-run-on sentences. 
We discard all other error annotations and combine the NUCLE train and CoNLL test sets, which we call \textit{RealESL}.

Only 1\% of sentences in RealESL are run-ons, which may not be the case in other forms of ESL corpora. So for a fair comparison we down-sample the run-on sentences in FakeESL to form a test set with the same distribution as RealESL, FakeESL-1\%.
Table~\ref{tab:3} describes the size of our data sets.

\begin{table}[t!]
\begin{center}
\small
\fontsize{7}{10}
\selectfont
\begin{tabular}{|l|l|l|l|l|}
\hline & Dataset & RO & Non-RO & Total \\ \hline
Train & FakeGiga-Train & 2.76M (61\%) & 1.75M (39\%) & 4.51M \\ \hline 
\multirow{4}{*}{Test} 
& FakeGiga & 28,232 (11\%) & 218,076 (89\%) & 246,308 \\ 
& RealESL  & 542 (1\%) & 58,987 (99\%) & 59,529 \\ 
& FakeESL  & 5,600 (9\%) & 56,350 (91\%) & 61,950\\
& FakeESL-1\%  & 560 (1\%) & 56,350 (99\%) & 56,910\\
\hline 
% & NUCLE & 483 & 56,350 & 56,833 \\ \cline{2-5}
% & CoNLL & 59 & 2,637 & 2,696 \\ \hline 
\end{tabular}
\end{center}
\caption{\label{tab:3} Number of run-on (RO) and non-run-on (Non-RO) sentences in our datasets.} % on test datasets we used for evaluation. ROs means Run-on sentences and Non-ROs means error free sentences. 
% ESL includes sentences from NUCLE train and the CoNLL-13 and CoNLL-14 shared-task test sets. }
\end{table}

\begin{table*}[!h!t]
\small
% \fontsize{10}{12}
% \selectfont
\begin{center}
\begin{tabular}{|l|lll|lll||lll|lll|}
\hline
& \multicolumn{6}{c||}{\textit{Clean v. Noisy - Artificial Data}}
& \multicolumn{6}{c|}{\textit{Real v. Artificial - Noisy Data}}\\
\cline{2-13}
& \multicolumn{3}{c|}{FakeGiga} & \multicolumn{3}{c||}{FakeESL}  & \multicolumn{3}{c}{RealESL} & \multicolumn{3}{|c|}{FakeESL-1\%} \\ \cline{2-13}
& $P$ & $R$ & \(F_{0.5}\) & $P$ & $R$ & \(F_{0.5}\) & $P$ & $R$ & \(F_{0.5}\) & $P$ & $R$ & \(F_{0.5}\) \\
\hline
Random & 0.10 & 0.10 & 0.10 
	& 0.09 & 0.09 & 0.09 
    & 0.01 & 0.01 & 0.01 
    & 0.01 & 0.01 & 0.01 \\
Punctuator-EU & 0.22 & 0.45 & 0.25
	& 0.74 &  0.48 &   \bf 0.67 
    & 0.11 & \bf 0.65 &   0.13 
    & 0.12 & \textbf{0.67} &  0.14\\
Punctuator-RO & 0.78 & 0.57 & 0.73 
	& 0.58 & \bf 0.51 & 0.56
	& 0.11 & 0.31 & 0.13 
    & 0.18 & 0.52 & 0.21 \\
roCRF & \bf 0.89 & 0.49 &   0.76  
	& \bf 0.83 &  0.24 &  0.55 
    & \bf 0.34 & 0.27 &  \bf 0.32
    & \bf 0.32 & 0.24 &  0.30 \\
roS2S & 0.84 & \bf 0.94  & \bf 0.86 
		& 0.77 & 0.44 &  \bf 0.67 
        & 0.30 & 0.32 &  0.31 
        & 0.30 & 0.34 &  \bf 0.31 \\
\hline
\end{tabular}
\end{center}
\caption{\label{tab:results-all} Performance on clean v. noisy artificial data with 10\% run-ons, and real v. artificial data with 1\% run-ons.}
\end{table*}

\section{Experiments}

\paragraph{Metrics:} We report precision, recall, and the $F_{0.5}$ score. In  GEC, precision is more important than recall, and therefore the standard metric for evaluation is $F_{0.5}$, which weights precision twice as much as recall.

\paragraph{Baselines:} We report results on a balanced random baseline and state-of-the-art models from whole-sentence GEC (NUS18) and punctuation restoration (the Punctuator).
NUS18 is the released GEC model of \citet{chollampatt2018multilayer}, trained on two GEC corpora, NUCLE and Lang-8~\cite{mizumoto-EtAl:2011:IJCNLP-2011}.
We test two versions of the Punctuator: \textit{Punctuator-EU} is the released model, trained on English Europarl~v7~\cite{koehn2005europarl}, and \textit{Punctuator-RO}, which we trained on artificial clean data (FakeGiga-Train) using the authors' code.\footnote{\url{github.com/ottokart/punctuator2}}
% CN:deleted ref to Punctuator since we already have it in nrelated work

% \citet[\textit{Punctuator}]{tilk16punctuator}. 
% We use the authors' pretrained models: NUS18 was trained on the GEC corpora Lang-8~\cite{mizumoto-EtAl:2011:IJCNLP-2011} and NUCLE. %\footnote{The reader will question our decision to use a model trained our test set, which we will address further in \S~\ref{sec:results}.}
% Punctuator was trained on English Europarl v7~\cite{koehn2005europarl} and we also retrained it on artificial data (Gigaword-60\%). % based entirely on textual features. 

\paragraph{roCRF:} We train our model with $\ell1$-regularization and $c=10$ using the CRF++ toolkit.\footnote{Version 0.59, \url{github.com/taku910/crfpp/}}
Only features that occur at least 5 times in the training set were included.
Spaces are labeled to contain missing punctuation when the marginal probability is less than 0.70. Parameters are tuned to $F_{0.5}$ on 25k held-out sentences.

\paragraph{roS2S:} Both the encoder and decoder have a single layer, 1028-dimensional hidden states, and a vocabulary of 100k words. 
We limit the input sequences to 100 words and use 300-dimensional pre-trained GloVe word embeddings~\cite{pennington-socher-manning:2014:EMNLP2014}.
The dropout rate is 0.5 and mini-batches are size 128. We train using Ada-grad with a learning rate of 0.0001 and a decay of 0.5. 
\\
% \linebreak
% \noindent Our CRF and Seq2Seq models are trained on artificial data from formal, grammatical text (Gigaword-60\% in Table~\ref{tab:3}).
% , and we evaluate on artificial data from clean (Gigaword) and noisy (NUCLE) sources. 
% We additionally test all models on naturally occurring run-ons from NUCLE. 

 %~\cref{tab:results-fake} lists the scores by test set.

\section{Results and Analysis}\label{sec:results}

Results are shown in Table~\ref{tab:results-all}. A correct judgment is where a run-on sentence is detected and a {\tt PERIOD} is inserted in the right place. 
Across all datasets, roCRF has the highest precision. We speculate that roCRF consistently has the highest precision because it is the only model to use POS and syntactic features, which may restrict the occurrence of false positives by identifying longer distance, structural dependencies.
roS2S is able to generalize better than roCRF, resulting in higher recall with only a moderate impact on precision. On all datasets except RealESL, roS2S consistently has the highest overall $F_{0.5}$ score.
In general, Punctuator has the highest recall, probably because it is trained for a more general purpose task and tries to predict punctuation at each possible position, resulting in lower precision than the other models.

NUS18 predicts only a few false positives and no true positives, so $P = R = 0$ and we exclude it from the results table.
Even though NUS18 is trained on NUCLE, which RealESL encompasses, its very poor performance is not too surprising given the infrequency of run-ons in NUCLE.

\paragraph{Clean v. Noisy} In the first set of experiments (columns 2 and 3), we compare models on clean text (FakeGiga), which has no other grammatical mistakes, and noisy text (FakeESL), which may have several other errors in each sentence.
Punctuator-EU is the only model which improves when tested on the noisy artificial data compared to the clean. It is possible that the speech transcripts used for training Punctuator-EU more closely resemble FakeESL, which is less formal than FakeGiga. All other models do worse, which could be due to overfitting FakeGiga.
However, further work is needed to determine how much of the performance drop is due to a domain mismatch versus the frequency of grammatical mistakes in the data. 

\paragraph{Real v. Artificial} So far, we have only used artificially generated data for training and testing.
The second set of experiments (columns 4 and 5) determines if it is easier to correct run-on sentences that are artificially generated compared to those that occur naturally. 
The Punctuators do poorly on this data because they are too liberal, evidenced by the high recall and very low precision.
Our models, roCRF and roS2S, outperform the Punctuators and have similar performance on both the real and artificial run-ons (RealESL and FakeESL-1\%). 
roCRF has significantly higher precision on RealESL while roS2S has significantly higher recall and F$_0.5$ on RealESL and FakeESL-1\% (with bootstrap resampling, $p < 0.05$).
This supports the use of artificially generated run-on sentences as training data for this task.

\section{Conclusions} 
% \todo{Let's leave this until the rest of the paper is sorted.}
% 1 . 
% The task of run-on sentence correction is one of the hardest among the grammatical error correction tasks, given the fact that it is related to not only syntactic information but also semantic information \todoc{I'd delete mention of semantics because we don't touch it at all in the rest of the paper} and that it is on sentence level. In this paper we introduced \todoc{``introduced'' too strong probably} the run-on sentence correction task by providing a huge artificial dataset developed from Gigaword and a real world test dataset developed from NUCLE and CoNLL-13 and CoNLL-14 dataset. We also provided two models based on sequence to sequence RNN and CRFs respectively. Both outperform one of the current state-of-the-art punctuation restoration system in this task. 

Correcting run-on sentences is a challenging task that has not been individually targeted in earlier GEC models.
% is more difficult than other grammatical errors due to the under-representation of this mistake in annotated GEC corpora. 
% We demonstrated how difficult this task is for leading GEC and punctuation restoration systems and 
We have developed two new models for run-on sentence correction: a syntax-aware CRF model, roCRF, and a Seq2Seq model, roS2S.
Both of these outperform leading models for punctuation restoration and grammatical error correction on this task.
In particular, roS2S has very strong performance, with $F_{0.5} = 0.86$ and $F_{0.5} = 0.67$ on run-ons generated from clean and noisy data, respectively.
roCRF has very high precision ($0.83 \leq P \leq 0.89$) but low recall, meaning that it does not generalize as well as the leading system, roS2S.

Run-on sentences have low frequency in annotated GEC data, so we experimented with artificially generated training data. 
We chose clean newswire text as the source for training data to ensure there were no unlabeled naturally occurring run-ons in the training data. Using ungrammatical text as a source of artificial data is an area of future work. 
The results of this study are inconclusive in terms of how much harder the task is on clean versus noisy text. 
However, our findings suggest that artificial run-ons are similar to naturally occurring run-ons in ungrammatical text because models trained on artificial data do just as well predicting real run-ons as artificial ones.

In this work, we found that a leading GEC model~\cite{chollampatt2018multilayer} does not correct any run-on sentences, even though there was an overlap between the test and training data for that model.
This supports the recent work of \citet{choshen-abend-acl2018}, who found that GEC systems tend to ignore less frequent errors due to reference bias. 
Based on our work with run-on sentences, a common error type that is infrequent in annotated data, we strongly encourage future GEC work to address low-coverage errors.

\section*{Acknowledgments}
We thank the three anonymous reviewers for their helpful feedback.

\bibliography{emnlp2018}

\begin{thebibliography}{30}
\expandafter\ifx\csname natexlab\endcsname\relax\def\natexlab#1{#1}\fi

\bibitem[{Bahdanau et~al.(2015)Bahdanau, Cho, and Bengio}]{bahdanau14neural}
Dzmitry Bahdanau, Kyunghyun Cho, and Yoshua Bengio. 2015.
\newblock Neural machine translation by jointly learning to align and
  translate.
\newblock In \emph{Proceedings of the Sixth International Conference on
  Learning Representations (ICLR)}.

\bibitem[{Che et~al.(2016)Che, Wang, Yang, and Meinel}]{che16punctuation}
Xiaoyin Che, Cheng Wang, Haojin Yang, and Christoph Meinel. 2016.
\newblock Punctuation prediction for unsegmented transcript based on word
  vector.
\newblock In \emph{Proceedings of the Tenth International Conference on
  Language Resources and Evaluation (LREC 2016)}, Paris, France. European
  Language Resources Association (ELRA).

\bibitem[{Chollampatt and Ng(2018)}]{chollampatt2018multilayer}
Shamil Chollampatt and Hwee~Tou Ng. 2018.
\newblock A multilayer convolutional encoder-decoder neural network for
  grammatical error correction.
\newblock In \emph{AAAI Conference on Artificial Intelligence}.

\bibitem[{Choshen and Abend(2018)}]{choshen-abend-acl2018}
Leshem Choshen and Omri Abend. 2018.
\newblock Inherent biases in reference-based evaluation for grammatical error
  correction.
\newblock In \emph{Proceedings of the 56th Annual Meeting of the Association
  for Computational Linguistics (Volume 1: Long Papers)}, pages 632--642.
  Association for Computational Linguistics.

\bibitem[{Dahlmeier and Ng(2011)}]{dahlmeier-ng:2011:ACL-HLT2011}
Daniel Dahlmeier and Hwee~Tou Ng. 2011.
\newblock Grammatical error correction with alternating structure optimization.
\newblock In \emph{Proceedings of the 49th Annual Meeting of the Association
  for Computational Linguistics: Human Language Technologies}, pages 915--923,
  Portland, Oregon, USA. Association for Computational Linguistics.

\bibitem[{Dahlmeier et~al.(2013)Dahlmeier, Ng, and
  Wu}]{dahlmeier-ng-wu:2013:BEA8}
Daniel Dahlmeier, Hwee~Tou Ng, and Siew~Mei Wu. 2013.
\newblock Building a large annotated corpus of learner english: The {NUS Corpus
  of Learner English}.
\newblock In \emph{Proceedings of the Eighth Workshop on Innovative Use of NLP
  for Building Educational Applications}, pages 22--31, Atlanta, Georgia.
  Association for Computational Linguistics.

\bibitem[{Felice et~al.(2014)Felice, Yuan, Andersen, Yannakoudakis, and
  Kochmar}]{felice-EtAl:2014:W14-17}
Mariano Felice, Zheng Yuan, {\O}istein~E. Andersen, Helen Yannakoudakis, and
  Ekaterina Kochmar. 2014.
\newblock Grammatical error correction using hybrid systems and type filtering.
\newblock In \emph{Proceedings of the Eighteenth Conference on Computational
  Natural Language Learning: Shared Task}, pages 15--24, Baltimore, Maryland.
  Association for Computational Linguistics.

\bibitem[{Gravano et~al.(2009)Gravano, Jansche, and
  Bacchiani}]{gravano2009restoring}
Agustin Gravano, Martin Jansche, and Michiel Bacchiani. 2009.
\newblock Restoring punctuation and capitalization in transcribed speech.
\newblock In \emph{Proceedings of the International Conference on Acoustics,
  Speech and Signal Processing (ICASSP)}, pages 4741--4744. IEEE.

\bibitem[{Israel et~al.(2012)Israel, Tetreault, and
  Chodorow}]{israel-tetreault-chodorow:2012:NAACL-HLT}
Ross Israel, Joel Tetreault, and Martin Chodorow. 2012.
\newblock Correcting comma errors in learner essays, and restoring commas in
  newswire text.
\newblock In \emph{Proceedings of the 2012 Conference of the North American
  Chapter of the Association for Computational Linguistics: Human Language
  Technologies}, pages 284--294, Montr\'{e}al, Canada. Association for
  Computational Linguistics.

\bibitem[{Junczys-Dowmunt and
  Grundkiewicz(2016)}]{junczysdowmunt-grundkiewicz:2016:EMNLP2016}
Marcin Junczys-Dowmunt and Roman Grundkiewicz. 2016.
\newblock Phrase-based machine translation is state-of-the-art for automatic
  grammatical error correction.
\newblock In \emph{Proceedings of the 2016 Conference on Empirical Methods in
  Natural Language Processing}, pages 1546--1556, Austin, Texas. Association
  for Computational Linguistics.

\bibitem[{Junczys-Dowmunt et~al.(2018)Junczys-Dowmunt, Grundkiewicz, Guha, and
  Heafield}]{junczysdowmunt-EtAl:2018:N18-1}
Marcin Junczys-Dowmunt, Roman Grundkiewicz, Shubha Guha, and Kenneth Heafield.
  2018.
\newblock Approaching neural grammatical error correction as a low-resource
  machine translation task.
\newblock In \emph{Proceedings of the 2018 Conference of the North American
  Chapter of the Association for Computational Linguistics: Human Language
  Technologies, Volume 1 (Long Papers)}, pages 595--606, New Orleans,
  Louisiana. Association for Computational Linguistics.

\bibitem[{Kagan(1980)}]{kagan1980run}
Dona~M Kagan. 1980.
\newblock Run-on and fragment sentences: An error analysis.
\newblock \emph{Research in the Teaching of English}, 14(2):127--138.

\bibitem[{Koehn(2005)}]{koehn2005europarl}
Philipp Koehn. 2005.
\newblock Europarl: A parallel corpus for statistical machine translation.
\newblock In \emph{MT summit}, volume~5, pages 79--86.

\bibitem[{Leacock et~al.(2014)Leacock, Chodorow, Gamon, and
  Tetreault}]{leacock2014automated}
Claudia Leacock, Martin Chodorow, Michael Gamon, and Joel Tetreault. 2014.
\newblock Automated grammatical error detection for language learners.
\newblock \emph{Synthesis lectures on human language technologies},
  7(1):1--170.

\bibitem[{Lee et~al.(2014)Lee, Yeung, and Chodorow}]{Lee-EtAl:2014:PACLIC}
John Lee, Chak~Yan Yeung, and Martin Chodorow. 2014.
\newblock Automatic detection of comma splices.
\newblock In \emph{Proceedings of the 28th Pacific Asia Conference on Language,
  Information, and Computation}, pages 551--560, Phuket,Thailand. Department of
  Linguistics, Chulalongkorn University.

\bibitem[{Lu and Ng(2010)}]{lu-ng:2010:EMNLP}
Wei Lu and Hwee~Tou Ng. 2010.
\newblock Better punctuation prediction with dynamic conditional random fields.
\newblock In \emph{Proceedings of the 2010 Conference on Empirical Methods in
  Natural Language Processing}, pages 177--186, Cambridge, MA. Association for
  Computational Linguistics.

\bibitem[{Madnani et~al.(2012)Madnani, Tetreault, and Chodorow}]{W12-2005}
Nitin Madnani, Joel Tetreault, and Martin Chodorow. 2012.
\newblock Exploring grammatical error correction with not-so-crummy machine
  translation.
\newblock In \emph{Proceedings of the Seventh Workshop on Building Educational
  Applications Using NLP}, pages 44--53. Association for Computational
  Linguistics.

\bibitem[{Mizumoto et~al.(2011)Mizumoto, Komachi, Nagata, and
  Matsumoto}]{mizumoto-EtAl:2011:IJCNLP-2011}
Tomoya Mizumoto, Mamoru Komachi, Masaaki Nagata, and Yuji Matsumoto. 2011.
\newblock Mining revision log of language learning sns for automated japanese
  error correction of second language learners.
\newblock In \emph{Proceedings of 5th International Joint Conference on Natural
  Language Processing}, pages 147--155, Chiang Mai, Thailand. Asian Federation
  of Natural Language Processing.

\bibitem[{Napoles et~al.(2012)Napoles, Gormley, and
  Van~Durme}]{napoles-gormley-vandurme:2012:AKBC-WEKEX}
Courtney Napoles, Matthew Gormley, and Benjamin Van~Durme. 2012.
\newblock Annotated {G}igaword.
\newblock In \emph{Proceedings of the Joint Workshop on Automatic Knowledge
  Base Construction and Web-scale Knowledge Extraction (AKBC-WEKEX)}, pages
  95--100, Montr{\'e}al, Canada. Association for Computational Linguistics.

\bibitem[{Ng et~al.(2014)Ng, Wu, Briscoe, Hadiwinoto, Susanto, and
  Bryant}]{ng-EtAl:2014:W14-17}
Hwee~Tou Ng, Siew~Mei Wu, Ted Briscoe, Christian Hadiwinoto, Raymond~Hendy
  Susanto, and Christopher Bryant. 2014.
\newblock The {CoNLL}-2014 shared task on grammatical error correction.
\newblock In \emph{Proceedings of the Eighteenth Conference on Computational
  Natural Language Learning: Shared Task}, pages 1--14, Baltimore, Maryland.
  Association for Computational Linguistics.

\bibitem[{Ng et~al.(2013)Ng, Wu, Wu, Hadiwinoto, and
  Tetreault}]{ng-EtAl:2013:CoNLLST}
Hwee~Tou Ng, Siew~Mei Wu, Yuanbin Wu, Christian Hadiwinoto, and Joel Tetreault.
  2013.
\newblock The {CoNLL}-2013 shared task on grammatical error correction.
\newblock In \emph{Proceedings of the Seventeenth Conference on Computational
  Natural Language Learning: Shared Task}, pages 1--12, Sofia, Bulgaria.
  Association for Computational Linguistics.

\bibitem[{Peitz et~al.(2011)Peitz, Freitag, Mauser, and
  Ney}]{peitz2011modeling}
Stephan Peitz, Markus Freitag, Arne Mauser, and Hermann Ney. 2011.
\newblock Modeling punctuation prediction as machine translation.
\newblock In \emph{International Workshop on Spoken Language Translation
  (IWSLT) 2011}.

\bibitem[{Pennington et~al.(2014)Pennington, Socher, and
  Manning}]{pennington-socher-manning:2014:EMNLP2014}
Jeffrey Pennington, Richard Socher, and Christopher Manning. 2014.
\newblock Glove: Global vectors for word representation.
\newblock In \emph{Proceedings of the 2014 Conference on Empirical Methods in
  Natural Language Processing (EMNLP)}, pages 1532--1543, Doha, Qatar.
  Association for Computational Linguistics.

\bibitem[{Rozovskaya and Roth(2011)}]{rozovskaya-roth:2011:ACL-HLT2011}
Alla Rozovskaya and Dan Roth. 2011.
\newblock Algorithm selection and model adaptation for {ESL} correction tasks.
\newblock In \emph{Proceedings of the 49th Annual Meeting of the Association
  for Computational Linguistics: Human Language Technologies}, pages 924--933,
  Portland, Oregon, USA. Association for Computational Linguistics.

\bibitem[{Tetreault et~al.(2010)Tetreault, Foster, and
  Chodorow}]{tetreault-foster-chodorow:2010:Short}
Joel Tetreault, Jennifer Foster, and Martin Chodorow. 2010.
\newblock Using parse features for preposition selection and error detection.
\newblock In \emph{Proceedings of the ACL 2010 Conference Short Papers}, pages
  353--358, Uppsala, Sweden. Association for Computational Linguistics.

\bibitem[{Tilk and Alum{\"a}e(2016)}]{tilk16punctuator}
Ottokar Tilk and Tanel Alum{\"a}e. 2016.
\newblock Bidirectional recurrent neural network with attention mechanism for
  punctuation restoration.
\newblock In \emph{Proceedings of Interspeech}, pages 3047--3051.

\bibitem[{Watcharapunyawong and Usaha(2012)}]{watcharapunyawong2012thai}
Somchai Watcharapunyawong and Siriluck Usaha. 2012.
\newblock Thai efl students’ writing errors in different text types: The
  interference of the first language.
\newblock \emph{English Language Teaching}, 6(1):67.

\bibitem[{Xie et~al.(2018)Xie, Genthial, Xie, Ng, and
  Jurafsky}]{xie-EtAl:2018:N18-1}
Ziang Xie, Guillaume Genthial, Stanley Xie, Andrew Ng, and Dan Jurafsky. 2018.
\newblock Noising and denoising natural language: Diverse backtranslation for
  grammar correction.
\newblock In \emph{Proceedings of the 2018 Conference of the North American
  Chapter of the Association for Computational Linguistics: Human Language
  Technologies, Volume 1 (Long Papers)}, pages 619--628, New Orleans,
  Louisiana. Association for Computational Linguistics.

\bibitem[{Yannakoudakis et~al.(2011)Yannakoudakis, Briscoe, and
  Medlock}]{yannakoudakis-briscoe-medlock:2011:ACL-HLT2011}
Helen Yannakoudakis, Ted Briscoe, and Ben Medlock. 2011.
\newblock A new dataset and method for automatically grading {ESOL} texts.
\newblock In \emph{Proceedings of the 49th Annual Meeting of the Association
  for Computational Linguistics: Human Language Technologies}, pages 180--189,
  Portland, Oregon, USA. Association for Computational Linguistics.

\bibitem[{Yuan and Briscoe(2016)}]{yuan-briscoe:2016:N16-1}
Zheng Yuan and Ted Briscoe. 2016.
\newblock Grammatical error correction using neural machine translation.
\newblock In \emph{Proceedings of the 2016 Conference of the North American
  Chapter of the Association for Computational Linguistics: Human Language
  Technologies}, pages 380--386, San Diego, California. Association for
  Computational Linguistics.

\end{thebibliography}
\bibliographystyle{acl_natbib_nourl}

\end{document}

\appendix

\section{Supplemental Material}
\label{sec:supplemental}

\subsection{CRF Features}

In the CRF described in \S~\ref{sec:crf}, a sentence with $n$ tokens, $<tok_0, tok_1, ..., tok_n>$ is represented by a sequence of spaces, where a space $s_{i,j}$ is preceded by  $tok_i$ and followed by $tok_j$.
Each space $s_{i,j}$ is represented the following features:
cap(i) indicates 

\begin{itemize}
\item percent of words following $s_{i,j}$ $\frac{N - j}{N}$
\item p(n-gram($i-k, i-1$), .) > p(n gram($i-k, i$)) for k = 1,2,3
\item tok_i, pos_i, 
\end{itemize}

courtney.napoles@ip-10-11-2-209:~$ head -3 /projects/run_ons/results-for-paper/crf/crf-gigaword-20180509/crfpp-NUCLE-CONLL.feats
we	PRP	True	0	26	0.0	BOS	lt	lt	lt	0	NP	VP	phrase/phrase	SPACE
do	VBP	False	1	25	0.0	IN	lt	lt	lt	1	VBP	RB	preterminal/preterminal	SPACE
not	RB	False	2	24	0.1	IN	gt	lt	lt	1	RB	VP	preterminal/phrase	SPACE